\documentclass[11pt,a4paper]{article}

\usepackage[hyperref]{acl2019}
\usepackage{times}
\usepackage{latexsym}
\usepackage{url}

\usepackage[ruled]{algorithm2e}
\usepackage{soul}
\usepackage[utf8]{inputenc}
\usepackage{graphicx}
\usepackage{amsmath}
\usepackage{booktabs}
\usepackage{tabularx}
\usepackage{bbm}
\usepackage{amsfonts}
\usepackage{multirow}
\usepackage{subfig}
\usepackage{mathrsfs}
\usepackage{amsthm}
\usepackage{natbib}
\usepackage{amsmath,bm}

\newcommand\blfootnote[1]{%
  \begingroup
  \renewcommand\thefootnote{}\footnote{#1}%
  \addtocounter{footnote}{-1}%
  \endgroup
}

\aclfinalcopy 
\def\aclpaperid{297} 

\title{Selection Bias Explorations and Debias Methods for\\ Natural Language Sentence Matching Datasets}

\author{  
  Guanhua Zhang$^{1,2*}$, Bing Bai$^{1*}$, Jian Liang$^{1}$, Kun Bai$^1$,\\ \textbf{Shiyu Chang$^3$, Mo Yu$^3$, Conghui Zhu$^2$, Tiejun Zhao$^2$}\\
  $^1$Cloud and Smart Industries Group, Tencent, China\\
  $^2$Harbin Institute of Technology, China\\
  $^3$MIT-IBM Watson AI Lab, IBM Research, USA\\
  \texttt{\{guanhzhang,icebai,joshualiang,kunbai\}@tencent.com},\\
  \texttt{shiyu.chang@ibm.com}, \texttt{yum@us.ibm.com},
  \texttt{\{chzhu,tjzhao\}@hit-mtlab.net}
}

\date{}

\begin{document}
\maketitle
\begin{abstract}
Natural Language Sentence Matching (NLSM) has gained substantial attention from both academics and the industry, and rich public datasets contribute a lot to this process.  However, biased datasets can also hurt the generalization performance of trained models and give untrustworthy evaluation results. For many NLSM datasets, the providers \emph{select} some pairs of sentences into the datasets, and this sampling procedure can easily bring unintended pattern, \emph{i.e.}, selection bias.  One example is the QuoraQP dataset, where some content-independent na\"ive features are unreasonably predictive. Such features are the reflection of the selection bias and termed as the ``\emph{leakage features}.'' In this paper, we investigate the problem of selection bias on six NLSM datasets and find that four out of them are significantly biased. We further propose a training and evaluation framework to alleviate the bias. Experimental results on QuoraQP suggest that the proposed framework can improve the generalization ability of trained models, and give more trustworthy evaluation results for real-world adoptions.\blfootnote{* Equal contributions from both authors. This work was done when Guanhua Zhang was an intern at Tencent.}
\end{abstract}

\section{Introduction}
\label{sec:intro}
Natural Language Sentence Matching~(NLSM) aims at comparing two sentences and identifying the relationships~\citep{wang2017bilateral}, and serves as the core of many NLP tasks such as question answering and information retrieval~\citep{wang2016sentence}. Natural Language Inference~(NLI)~\citep{bowman2015large} and Semantic Textual Similarity~(STS)~\citep{wang2016sentence} are both typical NLSM problems. A large number of publicly available datasets have benefited the research to a great extent~\citep{kim2018semantic,wang2017bilateral,tien2018sentence}, including QuoraQP\footnote{https://data.quora.com/First-Quora-Dataset-Release-Question-Pairs}, SNLI~\citep{bowman2015large}, SICK~\citep{marelli2014sick}, \emph{etc.} These datasets provide resources for both training and evaluation of different algorithms~\citep{torralba2011unbiased}.

\begin{figure}[!t]
\includegraphics[width=7.2cm]{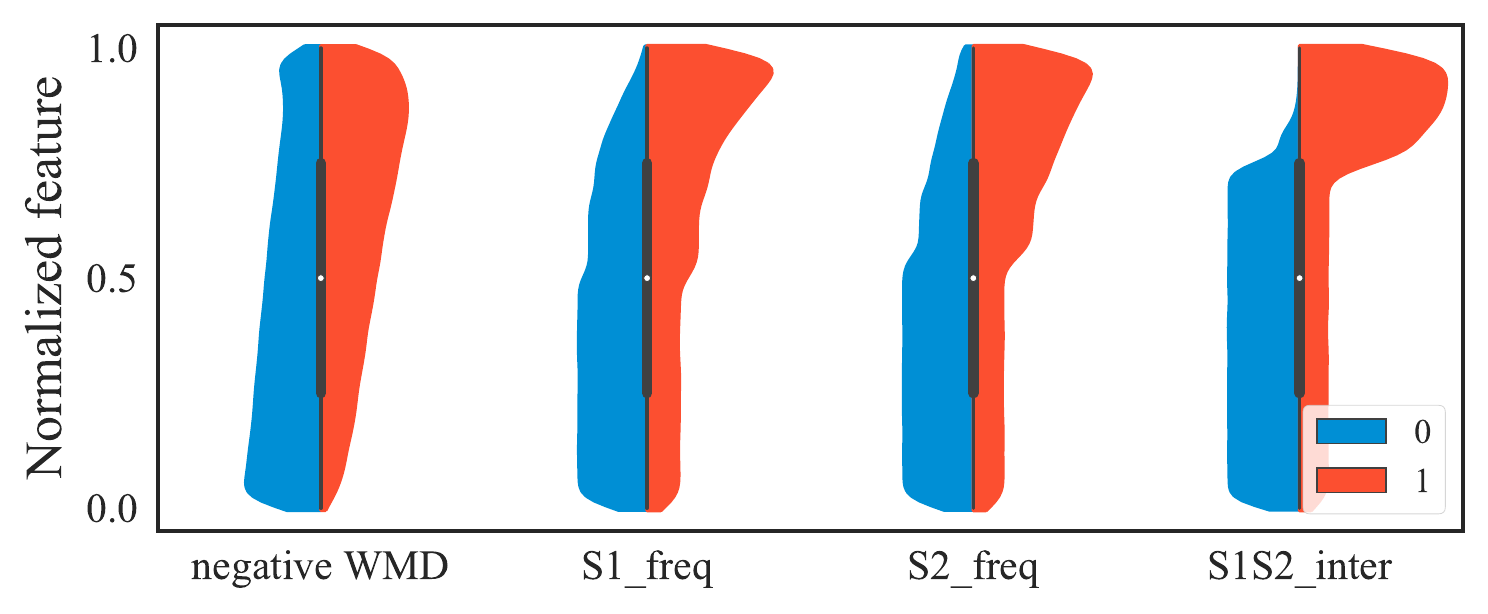}
\caption{Visualization for the distributions of normalized features versus the label in QuoraQP. The right part~(in red) represents the distributions of \texttt{duplicated} pairs, and the left part (in blue) represents the distributions of \texttt{not\_duplicated} pairs. Best viewed in color.}
\label{fig:leakage_vis}
\end{figure}

However, most of the datasets are prepared by conducting procedures involving a sampling process, which can easily introduce a \emph{selection bias}~\citep{heckman1977sample,zadrozny2004learning}. It would get even worse when the bias can reveal the label information, resulting in the ``\emph{leakage features},'' which are irrelevant to the content/semantic of the sentences but are predictive to the label. One example is the QuoraQP, a dataset on classifying whether two sentences are duplicated~(labeled as $1$) or not~(labeled as $0$), which has been widely used to evaluate STS models~\citep{gong2017natural,kim2018semantic,wang2017bilateral,devlin2018bert}. In QuoraQP, three leakage features have been identified, including \texttt{S1\_freq}, the number of occurrences of the first sentence in the dataset; \texttt{S2\_freq}, the number of occurrences of the second sentence; and \texttt{S1S2\_inter}, the number of sentences that are paired with both the first and the second sentences in the dataset for comparison.

\begin{figure}[!t]
\centerline{\includegraphics[width=5.5cm]{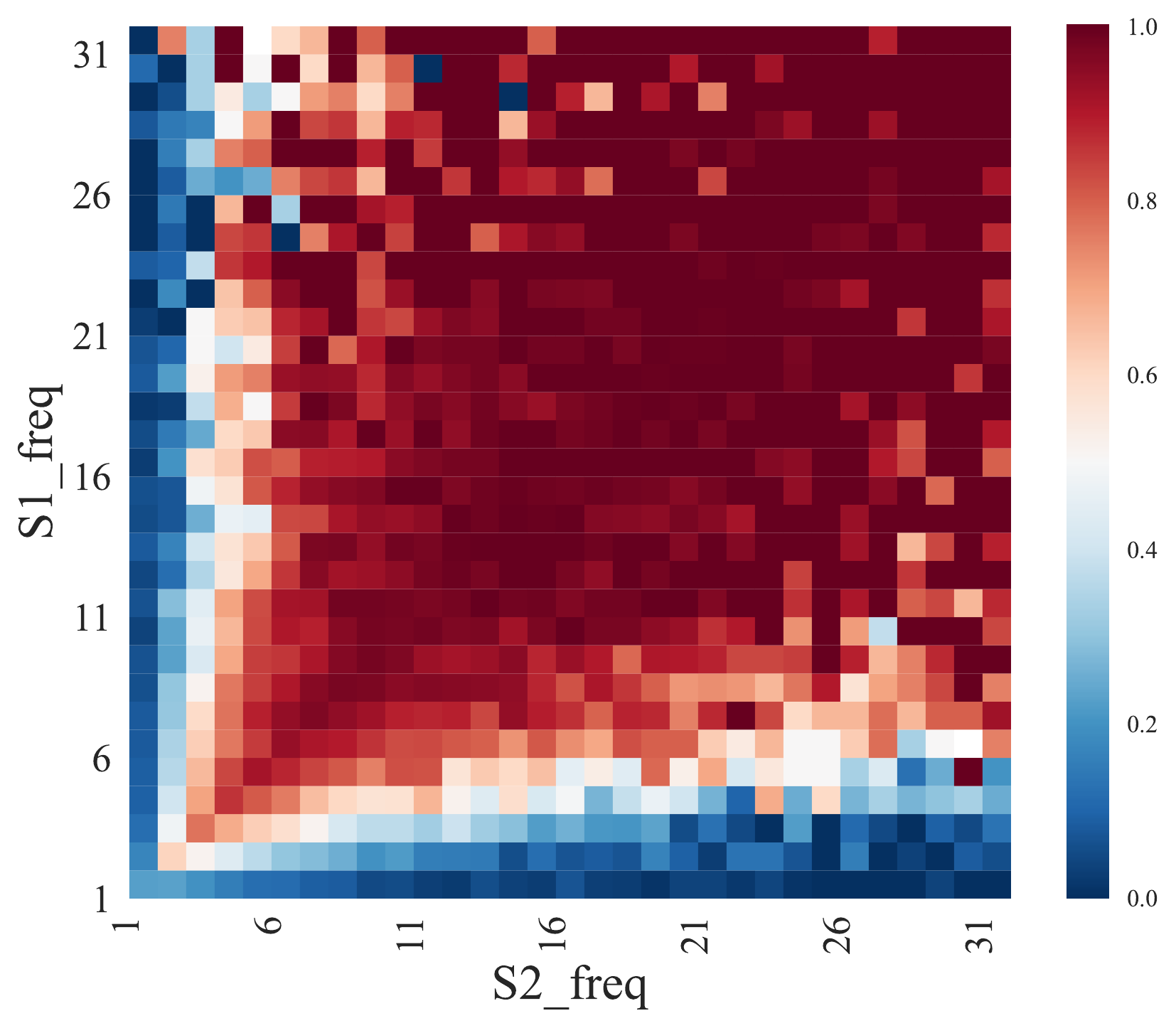}}
\caption{The averages of the labels under different \texttt{S1\_freq} and \texttt{S2\_freq}. Red squares indicate that the averages are close to 1, and blue squares indicate that the averages are close to 0. Best viewed in color.}
\label{fig:q1q2_vis}
\end{figure}

Figure~\ref{fig:leakage_vis} shows the distributions of normalized (negative) Word Mover's Distance~(WMD)~\citep{kusner2015word} and normalized leakage features versus the labels in QuoraQP. The features are all normalized to their quantiles. As illustrated, the leakage features are more predictive than the WMD, as the differences between the distributions of positive and negative pairs are more significant. Moreover, combining \texttt{S1\_freq} and \texttt{S2\_freq} can make even more accurate predictions as illustrated in Figure~\ref{fig:q1q2_vis}, where we calculate the averages of the labels under different \texttt{S1\_freq} and \texttt{S2\_freq}. We find that when both features' values are large, the pairs tend to be \texttt{duplicated}~(marked in red), while when one is large and the other is small, the pairs tend to be \texttt{not\_duplicated}~(marked in blue). 

These leakage features play a critical role in the QuoraQP competition\footnote{\url{https://www.kaggle.com/c/quora-question-pairs/discussion/34355} and \url{https://www.kaggle.com/c/quora-question-pairs/discussion/33168}}. As the evaluations are conducted with the same biased datasets, models that fit the bias pattern can take additional advantages over unbiased models, making the benchmark results untrustworthy. On the other hand, the bias pattern doesn't exist in the real-world, so if a model fits the bias pattern (intentionally or unintentionally), the generalization performance will be hurt, limiting the values of these datasets for further applications~\cite{torralba2011unbiased}.

In this paper, we study this problem and demonstrate the impact of the selection bias by a series of experiments. We focus on the selection bias embodied in the comparing relationships of sentences, and the main contributions of this paper are the answers to the following questions:
\begin{itemize}\setlength{\itemsep}{0pt}
    \item \textbf{Does selection bias exist in other NLSM datasets?} We identify four out of six publicly available datasets that suffer from the selection bias.
    \item \textbf{Would Deep Neural Network~(DNN)-based methods learn from the bias pattern unintentionally?} We find that Siamese-LSTM models trained on QuoraQP do capture the bias pattern.
    \item \textbf{Can we help the model learn the useful semantic pattern from the content without fitting the bias pattern?} We propose an easy-adopting method to mitigate the bias. Experiments show that this method can improve the generalization performance of the trained models.
    \item \textbf{Can we build an evaluation framework that gives us more reliable results for real-world adoption?} We propose a more trustworthy evaluation method that demonstrates consistent results with unbiased cross-dataset evaluations.
\end{itemize}

The rest of the paper is organized as follows. Section~\ref{sec:look_at_leakage} gives an empirical look at the selection bias on a variety of NLSM datasets and analyzes why the leakage features are effective. Section~\ref{sec:fit_bias} examines whether DNN-based methods fit the bias pattern unintentionally. Section~\ref{sec:unbiased_learning} introduces the training and evaluation framework to alleviate the biasedness. Taking QuoraQP as an example, we report the experimental results in Section~\ref{sec:experiments}. Section~\ref{sec:related_work} summarizes related work, and Section~\ref{sec:conclusion} draws the conclusion.

\section{Empirical Study of the Selection Bias}
\label{sec:look_at_leakage}

In this section, we investigate the problem of selection bias on six NLSM datasets and then analyze why the leakage features are effective. 

\subsection{Quantifying the Biasedness in Datasets}
\label{sec:leakage_in_datasets}

\begin{table*}[t!]
\noindent\makebox[\textwidth]{
\resizebox{\textwidth}{!}{
\begin{tabular}{c|c|c|c|c|c|c|c|c}
\hline
\multirow{2}*{\bf Method} & \bf \multirow{2}*{\bf SNLI} & \multicolumn{2}{c|}{\bf MultiNLI~} & \multirow{2}*{\bf QuoraQP} & \bf \bf \multirow{2}*{\bf MSRP} & \multicolumn{2}{c|}{\bf SICK} & \bf \multirow{2}*{\bf ByteDance} \\
 && \bf \small Matched & \bf \small Mismatched & && \bf \small NLI & \bf \small STS & \\
\hline
Majority      & 33.7 & 35.6 & 36.5 & 50.00 & 66.5 & 56.7 & 50.3 & 68.59 \\
Unlexicalized & 47.7 & 44.9 & 45.5 & 68.20 & 73.9 & 70.1 & 70.2 & 75.23 \\
LSTM          & 77.6$^*$ & 66.9$^\dagger$ & 66.9$^\dagger$ & 82.58$^\ddagger$ & 70.6$^\diamond$     & 71.3$^\top$      &70.2       & 86.45 \\
\hline
Leakage       & 36.6 & 32.1 & 31.1 & 79.63 & 66.7 & 56.7 & 55.5 & 78.24 \\
Advanced      & 39.1 & 32.7 & 33.8 & 80.47 & 67.9 & 57.5 & 56.3 & 85.73 \\
\hline
\hline
Leakage \emph{vs} Majority  & +8.61& -9.83 & -14.79 & +59.26 & +0.30 & 0.00& +10.34 & +14.07 \\
Advanced \emph{vs} Majority & +16.02 & -8.15 & -7.40 & +60.94 & +2.11 & +1.41 & +11.93 & +24.99 \\
\hline
\end{tabular}
}
}
\caption{\label{table:leakage_pred} The accuracy scores of predicting the label with unlexicalized features, leakage features, and advanced graph-based features and the relative improvements. Result with $^*$ is from~\citet{bowman2015large}. Results with $^\dagger$ are from~\citet{williams2018broad}. Result with $^\ddagger$ is from~\citet{wang2017bilateral}. Result with $^\diamond$ is from~\citet{shen2018baseline}. Result with $^\top$ is from~\citet{baudivs2016sentence}. Other results are based on our implementations. ``\%'' is omitted.}
\end{table*}

To quantify the severity of the leakage from the selection bias, we formulate a toy problem for NLSM. We predict the \emph{semantic relationship} of two sentences based on the \emph{comparing relationships} between sentences. We refer \emph{semantic relationship} of two sentences as their labels, for example, \texttt{duplicated} for STS and \texttt{entailment} for NLI, and \emph{comparing relationship} as whether they are paired for comparison in the dataset. Here we only consider the index of each sentence, and the actual content is not used. The formal problem definition is as follow:

\newtheorem{problem}{Problem}
\begin{problem}[\label{problem:link_pred} Leveraging the Leakage for NLSM]
Given a set of sentence ids $\mathbb{S}$, and a set of comparing relationships of the sentences $\mathbb{C}=\{\langle s_i, s_j \rangle \}, s_i, s_j \in \mathbb{S}$. The goal is to infer the semantic relationship between given pairs of sentence ids from $\mathbb{S}$.
\end{problem}

\begin{figure}[!t]
\centerline{\includegraphics[width=7.2cm]{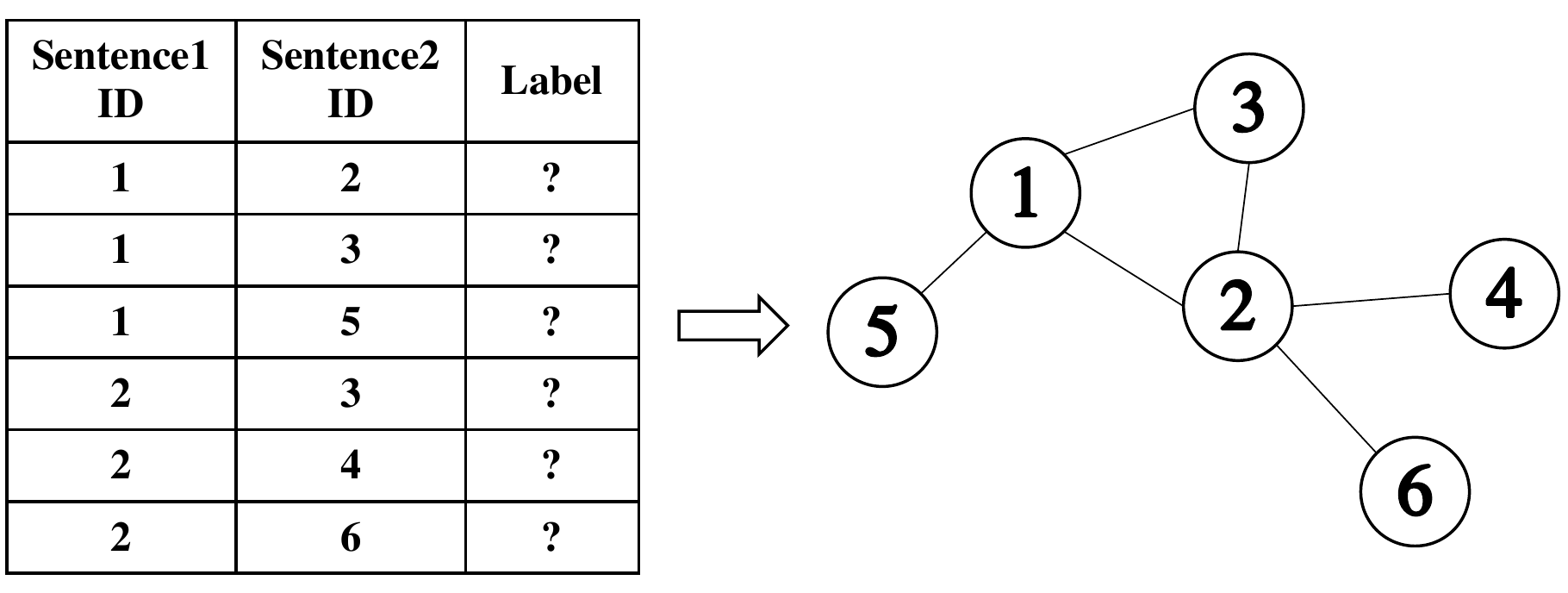}}
\caption{Illustration of the graph built for Problem~\ref{problem:link_pred}. We only use the comparing relationships to build the graph.}
\label{fig:toy_problem}
\end{figure}

This toy problem is indeed an edge classification problem \cite{Aggarwal2016Edge}, as we can construct a graph using the comparing relationships as illustrated in Figure~\ref{fig:toy_problem}. In addition, from the graph perspective, \texttt{S1\_freq} and \texttt{S2\_freq} are the degrees of nodes, and \texttt{S1S2\_inter} is the number of 2-hop paths connecting two nodes. Learning on the graph for this toy problem follows a transductive setting~\citep{ji2010graph}, where the graph is built with the comparing relationships of all the examples.

Based on the new problem definition, we investigate six NLSM datasets, including SNLI, MultiNLI~\citep{williams2018broad}, QuoraQP, MSRP~\citep{dolan2004unsupervised}, SICK and ByteDance\footnote{https://www.kaggle.com/c/fake-news-pair-classification-challenge}. We apply two different methods to classify the edges on the graph, including \textbf{Leakage} which uses the three leakage features introduced in Section~\ref{sec:intro} and \textbf{Advanced} which uses some more advanced graph-based features~\citep{perozzi2014deepwalk,zhou2009predicting,liben2007link} together with the three leakage features\footnote{The features are selected carefully to describe the local structure between two nodes and to prevent the model from remembering the exact ID of sentences to make inferences.}. We also report the results of three baselines, including \textbf{Majority} which predicts the most frequent label, \textbf{Unlexicalized} which uses 15 handcrafted features from the content of sentences~\citep{bowman2015large}~(\emph{e.g.}, the BLEU score~\citep{papineni2002bleu} of both sentences, the length difference between the two sentences, the percentage of overlap words, and so on) and \textbf{LSTM} which is a DNN-based method using sequences of word embeddings. All classifiers are Random Forests if no specific configuration is mentioned. The classifiers are trained with the training set, and we report the results on the testing set. More detailed settings are introduced in Appendix~\ref{sec:appendix}. The results are reported in Table~\ref{table:leakage_pred}.

Predicting semantic relationships without using sentence contents seems impossible. However, we find that the graph-based features~(\emph{Leakage} and \emph{Advanced}) make the problem feasible on a wide range of datasets. Specifically, on the datasets like QuoraQP and ByteDance, the leakage features are even more effective than the unlexicalized features.  One exception is that on MultiNLI, \emph{Majority} outperforms \emph{Leakage} and \emph{Advanced} significantly. Another interesting finding is that on SNLI and ByteDance, advanced graph-based features improve a lot over the leakage features, while on QuoraQP, the difference is very small.  Among all the tested datasets, only MSRP and SICK$_{\text{NLI}}$ are almost neutral to the leakage features. Note that their sizes are relatively small with only less than 10k samples. Results in Table \ref{table:leakage_pred} raise concerns about the impact of selection bias on the models and evaluation results.

\subsection{Why are the Leakage Features Effective?}
\label{sec:why_leakage}

As discussed in Section~\ref{sec:intro}, the leakage features are the reflection of selection bias.  Intuitively, if we construct a dataset for NLSM by randomly sampling some pairs of sentences, the resulting dataset would be extremely imbalanced, where the most of the pairs are \texttt{neutral} for NLI or \texttt{not\_duplicated} for STS. Thus, to make the dataset relatively balanced, a sampling strategy is often required. If the strategy is not well-designed, it will introduce a bias pattern into the dataset, which can be revealed by leakage features. Here we try to figure out why the leakage features are effective in aforementioned datasets. Since we do not have every detail about how they are constructed, we only analyze based on SNLI and QuoraQP.

\begin{figure}[!t]
\includegraphics[width=6.5cm]{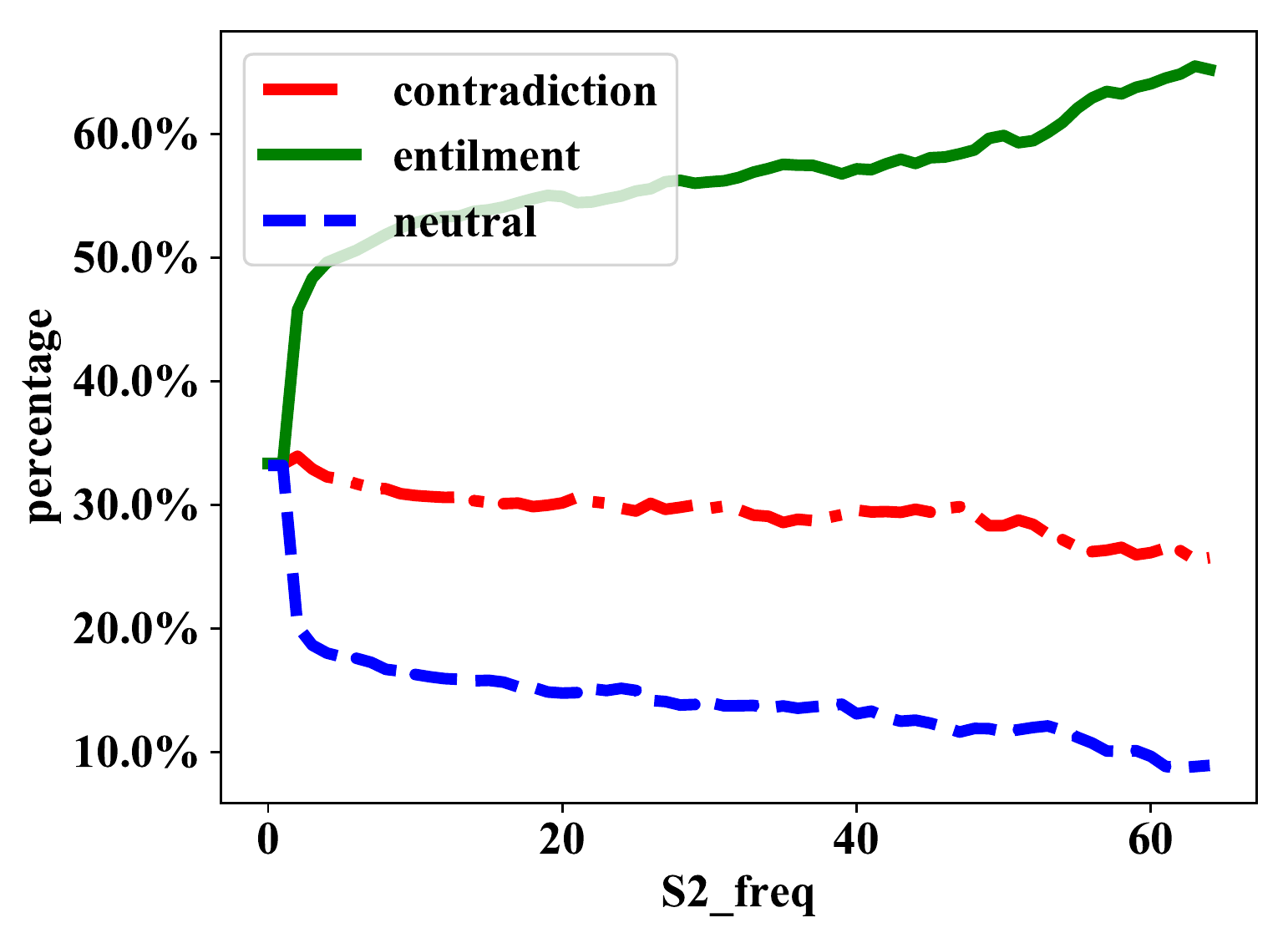}
\caption{The percentage of each label versus \texttt{S2\_freq} in SNLI.}
\label{fig:snli_feature2}
\end{figure}

During the preparation of SNLI, as introduced in~\citep{bowman2015large}, human workers are presented with ``premise scene descriptions,'' and asked to supply ``hypotheses'' for each of the three labels~(\emph{i.e.}, \texttt{entailment}, \texttt{neutral} and \texttt{contradiction}). However, it is found that some workers are ``\emph{reusing the same sentence for many different prompts},'' which might cause SNLI to suffer from selection bias. To validate, we calculate the percentage of each label versus \texttt{S2\_freq}, and the results are shown in Figure~\ref{fig:snli_feature2}. We see that the percentages of the three labels are similar when \texttt{S2\_freq} is small, but as \texttt{S2\_freq} increases, the label is more likely to be an \texttt{entailment}.

For QuoraQP dataset, the providers state that ``\emph{Our original sampling method returned an imbalanced dataset with many more true examples of duplicate pairs than non-duplicates. Therefore, we supplemented the dataset with negative examples. One source of negative examples were pairs of ``related questions'' which, although pertaining to similar topics, are not truly semantically equivalent.}'' Our hypothesis is that the way in which negative samples were supplemented is the reason why QuoraQP is so biased. For example, the newly added sentences of ``related questions'' may appear in the dataset for limited times, thus we get the phenomenon in Figure~\ref{fig:q1q2_vis}, \emph{i.e.}, if two sentences both appear for many times, the pair is likely to be \texttt{duplicated}, while if one of them appears for only a few times, the pair is likely to be \texttt{not\_duplicated}.

\begin{table}[t!]
\begin{center}
\resizebox{0.485\textwidth}{!}{
\begin{tabular}{c|c|c|c|c}
\hline
\bf Features & \bf \bf SNLI & \bf QuoraQP & \bf SICK$_{\text{STS}}$ & \bf ByteDance \\
\hline
\texttt{S1\_freq}                   & 33.7 & 65.90 & 54.5 & 68.61 \\
\texttt{S2\_freq}                   & 36.6 & 69.84 & 52.5 & 73.03\\
\texttt{S1S2\_inter}                & 33.7 & 79.66 & 50.8 & 76.63\\
$\urcorner$ \texttt{S1\_freq}       & 36.6 & 79.62 & 53.5 & 77.17\\
$\urcorner$ \texttt{S2\_freq}       & 33.7 & 79.66 & 53.0 & 77.44\\
$\urcorner$ \texttt{S1S2\_inter}    & 36.6 & 74.75 & 54.2 & 74.39\\
all                                 & 36.6 & 79.63 & 55.5 & 78.24 \\
\hline
Majority & 33.7 & 50.00 & 50.3 & 68.59\\
\hline
\end{tabular}
}
\end{center}
\caption{\label{table:ablantion} Ablation experiments of the three leakage features on the datasets. ``$\urcorner$'' means without the feature. We report the accuracy scores and ``\%'' is omitted.}
\end{table}

We conduct ablation experiments on the datasets where the leakage features are effective, \emph{i.e.}, SNLI, QuoraQP, SICK$_{\text{STS}}$ and ByteDance. The results are reported in Table~\ref{table:ablantion}. We can see that \texttt{S2\_freq} is more effective in SNLI, and \texttt{S1\_freq} plays a more critical role in SICK$_{\text{STS}}$, while in QuoraQP and ByteDance, \texttt{S1S2\_inter} is the most predictive.

Based on the experiments and observations, we conclude that existing datasets incline to be biased due to various reasons. More information about dataset preparations and further study are required to understand the problem and prevent bias from being introduced into future datasets.

\section{Do NN Models Fit the Bias Pattern Unintentionally?}
\label{sec:fit_bias}

In this section, we investigate whether DNN models are unintentionally fitting the bias pattern in addition to the semantic pattern. We train a classical Siamese-LSTM model\footnote{The detailed setting for the model is introduced in Section~\ref{sec:settings}} with the training set of QuoraQP, and make predictions on a synthetic dataset. Interestingly, we find that the results are significantly influenced by the bias pattern.

The synthetic dataset is built in the following way. We extract the distinct sentences from the training set of QuoraQP, then compare the sentences with themselves, finally we obtain 517,970 pairs in total. Since the two sentences in the pairs are identical, the labels are all \texttt{duplicated}. All three leakage features are the same, \emph{i.e.}, the numbers of occurrences of the sentence in the dataset. If the model can perfectly learn the semantic relationships between sentences, the predictions should be substantially the same for all the pairs.

\begin{figure}[!t]
\includegraphics[width=7.2cm]{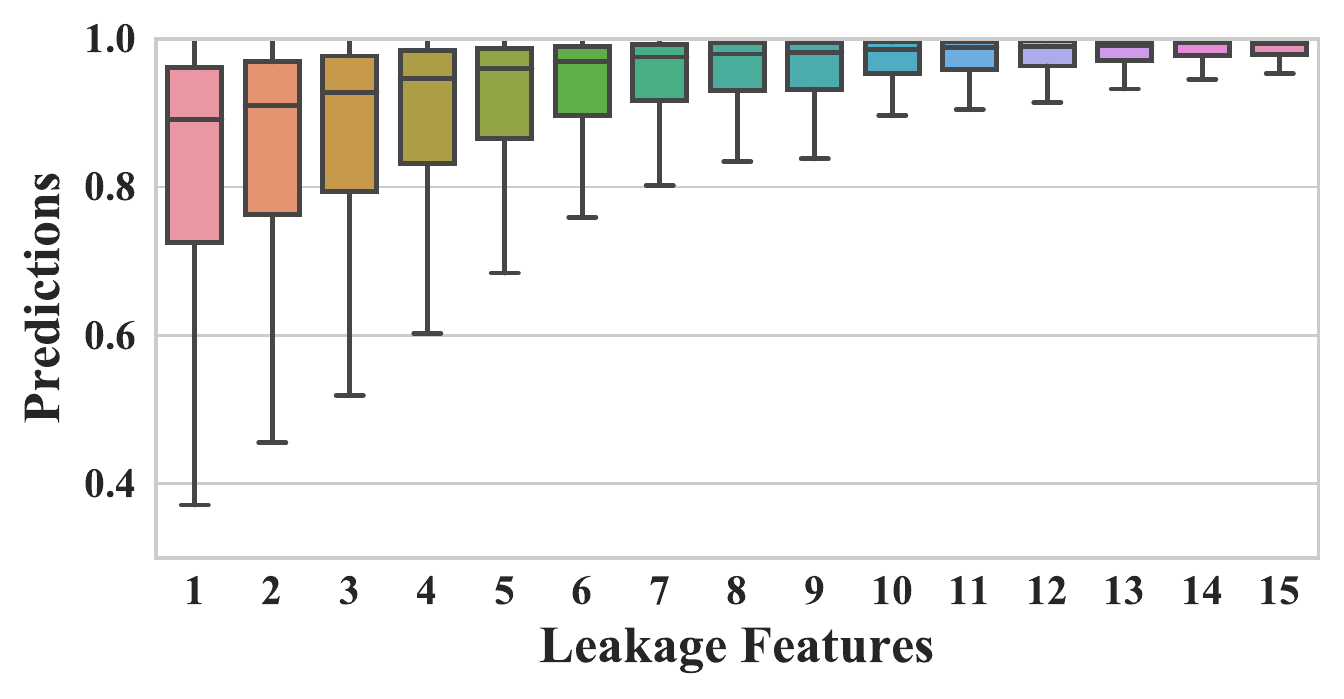}
\caption{Visualization of predicted scores versus the leakage feature. The boxes represent the upper quartiles to the lower quartiles of predicted scores, and the lowest datum is the 1.5 IQR of the lower quartile.}
\label{fig:duplicated_preds}
\end{figure}

To illustrate the predicted scores of duplication, we visualize them versus the leakage features in Figure~\ref{fig:duplicated_preds}, and the boxplot follows the Tukey boxplot style~\citep{frigge1989some}. Intriguingly, we find that even though the sentences in pairs are all identical, the model still tends to give lower scores of duplication to the pairs with leakage features equal to $1$. This result is consistent with the bias pattern shown in Figure~\ref{fig:q1q2_vis}, \emph{i.e.}, the data points in the bottom left corner tend to be \texttt{not\_duplicated}, compared with the data points in the top right corner which represent larger values of \texttt{S1\_freq} and \texttt{S2\_freq}. 

The results indicate that the model is unintentionally capturing the undesired bias pattern that only exists in the particular dataset. This will make an adverse effect on the generalization performance of the trained models~(to be illustrated in Section~\ref{sec:experimental_results}).

\section{Leakage-Neutral Learning and Evaluation Method}
\label{sec:unbiased_learning}

Given a biased dataset, can we eliminate the bias to train completely unbiased models? Unfortunately, this is very difficult due to that the bias is related with the labels, and we cannot have access to the labels of unselected samples~\cite{zadrozny2004learning}. In this paper, we propose to take a step back and define a \emph{leakage-neutral} distribution, which is more close to the real-world than the biased one. We make a few reasonable assumptions about it and how the biased dataset is generated from it. We demonstrate that we can train and evaluate models unbiased to the leakage-neutral distribution, with only the biased dataset.

\paragraph{Generation of the biased dataset from leakage-neutral distribution}
Assuming that there is a \emph{leakage-neutral} distribution $\mathscr{D}$ with domain $\mathcal{X} \times \mathcal{Y} \times \mathcal{L} \times \mathcal{S}$ where $\mathcal{X}$ is the semantic feature space, $\mathcal{Y}$ is the (binary) semantic label space, $\mathcal{L}$ is the sampling strategy feature space and $\mathcal{S}$ is the (binary) sampling intention space. The sampling intentions represent whether dataset providers want to select a positive sample or a negative sample. For example, $S=1$ means that the providers want to select a positive sample here.

We assume that samples $(x, y, l, s)$ are drawn independently from $\mathscr{D}$, then if $s=y$~(the label matches the sampling intention), the samples are selected into the dataset, otherwise, the samples are discarded. This operation results in the biased distribution $\widehat{\mathscr{D}}$ that are observed from the dataset.

In this section, we use uppercase letters, such as $Y$ and $S$, to represent random variables, and lowercase letters, such as $y$ and $s$, to represent specific values for samples. We use $P_{\widehat{\mathscr{D}}}(\cdot)$ to represent the probability on $\widehat{\mathscr{D}}$ and omit the subscripts for $\mathscr{D}$.

\paragraph{Assumptions about the leakage-neutral distribution}
We make the following assumptions about $\mathscr{D}$. The first one is the \emph{leakage-neutral} assumption defined as follows,
\begin{displaymath}
P(Y|L) = P(Y) \text{,}
\end{displaymath}
which means that the sampling strategy is independent with the labels, making the leakage-neutral distribution more close to the real-world.

The second one is that, given $L$, $S$ is independent with $X$ and $Y$ defined as follows,
\begin{displaymath}
P(S|X,Y,L) = P(S|L) \text{,}
\end{displaymath}
which means that the sampling strategy features can completely control the sampling intentions.

\paragraph{Leakage-neutral learning and evaluation method}

Based on the assumptions above, given a biased dataset, the proposed method works in the following way. 

Firstly, we estimate $P_{\widehat{\mathscr{D}}}(Y=1|l)$ from the dataset for all samples. In practice, this can be achieved by training classifiers and making cross-predictions. Since we don't have access to the true sampling strategy features, we use the leakage features from the graph instead, as they are the reflection of the biased sampling strategy.

Then we can get $P(S=1|l)$, the conditional probability of the sampling intention $S$ on $\mathscr{D}$ given $l$, using the following equation with $P(Y=1)$ given.

\begin{small}
\begin{equation}
\label{eq}
\begin{split}
P(&S=1|l) \\
& =\frac{P(Y=0)P_{\widehat{\mathscr{D}}}(Y=1|l)}{P(Y=0)P_{\widehat{\mathscr{D}}}(Y=1|l)+P(Y=1)P_{\widehat{\mathscr{D}}}(Y=0|l)}\text{.}
\end{split}
\end{equation}
\end{small}
The derivation of Equation~(\ref{eq}) is presented in Appendix~\ref{sec:proof2}.

Afterwards, we use $w=\frac{1}{P(S=y|l)}$ as the weights for the samples~(note that the labels $y$ are needed here). Training and evaluating with the weights can give us the results unbiased to the leakage-neutral distribution.

The step-by-step procedure for leakage-neutral learning and evaluation is presented in Algorithm~1. Note that our analyses and the proposed method are general enough for a variety of bias, as long as a sampling strategy feature is given, and can be easily extended to multi-class classification problems.

\begin{table}[t!]
\begin{center}
\resizebox{0.485\textwidth}{!}{
\begin{tabular}{p{0.15cm}<{\centering} p{8.55cm}}
\hline
\multicolumn{2}{l}{\textbf{Algorithm 1: Leakage-neutral Training and Evaluation}} \\
\hline
\multicolumn{2}{p{9.2cm}}{\textbf{Input}: The dataset $\{x, y\}$, the number of fold $K$ for cross prediction, and the prior probability $P(Y=1)$.} \\
\multicolumn{2}{l}{\textbf{Procedure}:}\\
01 & Extract the leakage features $l$ from the dataset.\\
02 & Estimate $P_{\widehat{\mathscr{D}}}(Y=1|l)$ for all samples by training classifiers and using $K$-fold cross-predicting strategy.\\
03 & Calculate $P(S=1|l)$ for all samples according to Equation~(\ref{eq}). \\
04 & Obtain the weights $w=\frac{1}{P(S=y|l)}$ for all samples and normalize the mean of the weights. \\
05 & Train and validate models with the training set and validation set respectively using $w$ as the sample weights.\\
06 & Evaluate the models with the testing set using $w$ as the sample weights.\\
\hline
\end{tabular}
}
\end{center}
\end{table}

\paragraph{Theoretical guarantee of unbiasedness}

Assuming that we know $P(S=y|l)$, and they are greater than zero for any $l$, the following theorem shows that we can obtain the loss unbiased to the leakage neutral distribution after using the sample weights.

\newtheorem{theorem}{Theorem}
\begin{theorem}[Unbiased Expectation]
\label{theorem:unbiased}
For any classifier $f=f(x, l)$, and for any loss function $\Delta = \Delta(f(x,l), y)$, if we use $w=\frac{P(S=Y)}{P(S=y|l)}$ as weights, then
\begin{small}
\begin{displaymath}
E_{x,y,l \sim \widehat{\mathscr{D}}} \Big[w\Delta\big(f(x,l), y\big) \Big] = E_{x, y, l \sim \mathscr{D}} \Big[ \Delta(f(x,l), y) \Big] \text{.}
\end{displaymath}
\end{small}
\end{theorem}

The proof is presented in Appendix~\ref{sec:proof1}. Since $P(S=Y)$ is only a number which does not affect the models, we can concentrate on the denominator, \emph{i.e.}, $P(S=y|l)$ and use $w=\frac{1}{P(S=y|l)}$ as the weights instead. The loss can be used for both training and evaluation unbiased to the leakage neutral distribution.

\section{Experimental Results for the Leakage-neutral Method on QuoraQP}
\label{sec:experiments}

In this section, we present the experimental results for leakage-neutral learning on QuoraQP. We demonstrate that the proposed learning framework can mitigate the bias and improve the generalization performance of trained models. Besides, the corresponding evaluation method can serve as a more reliable in-domain benchmark compared with the biased one. 

\subsection{Dataset Information and Weight Generation}

We use QuoraQP as our experimental dataset. We use the same dataset partition as~\citep{wang2017bilateral}.

We use the three leakage features for generating the weights. We use Random Forest classifiers to estimate $P_{\widehat{\mathscr{D}}}(Y=1|l)$, and the 100-fold cross predictions as the estimated values. $P(Y=1)$ is chosen to keep the proportion of the weights of positive and negative samples unchanged in order to prevent the influence of prior probabilities, and the mean of the weights is normalized to 1. The minimum weight of all the samples is $0.51$, and the maximum weight is $4953.17$.

\subsection{Experiment Settings}
\label{sec:settings}

We implement a classical Siamese-LSTM model with Keras and Tensorflow~\citep{abadi2016tensorflow} backend. Sequences of the embeddings of word tokens are fed into the LSTM layer with a hidden size of 128. Then the representations of both sentences, as well as the dot-production of the representations, go through a two Layer MLP where Batch Normalization~\citep{ioffe2015batch} is applied after every hidden layer. Dropout~\citep{srivastava2014dropout} with rate 0.5 is applied after the last hidden layer. We use the RMSProp~\citep{tieleman2012lecture} optimizer to train all the parameters. The learning rate starts at 1e-3, and decays at a fixed rate of 0.2 when performance does not improve on the validation set. We also use a gradient clipping of 5.0. The batch size is set to 256. All the results reported in this section are the average numbers of ten runs using the same hyper-parameters with different random initializations. Our implementation achieves slightly better performance compared with the results of the original Siamese-LSTM from~\citet{wang2017bilateral}.

We initialize our word embeddings with pre-trained GloVe 840B 300D vectors~\citep{pennington2014glove}, and the embeddings are kept fixed during training. All the sentences are cut off to have a maximum of 35 word tokens.

Note that the scale of weights of the different samples varies greatly. To prevent the model from jiggling during the mini-batch training, we use a sampling strategy for model training, \emph{i.e.}, we sample examples with probabilities proportional to the weights to get the data for every mini-batch\footnote{Codes and weights are published at \url{https://github.com/arthua196/Leakage-Neutral-Learning-for-QuoraQP}}.

\subsection{Evaluation Scheme}

To evaluate the effectiveness of leakage-neutral learning, we use the following strategy in our experiments. Firstly, we train and validate a model using the data from QuoraQP \emph{without} any weights. The model is referred to as \textbf{Biased Model}. Then we train and validate a model using the data from QuoraQP \emph{with} the weights, and the model is referred to as \textbf{Debiased Model}. These two models are evaluated with the following methods.

\begin{itemize}\setlength{\itemsep}{0pt}
    \item \textbf{Testing set evaluation}. We evaluate the models with the testing set of QuoraQP. Evaluation without the weights is named as \textbf{Biased Eva}, and evaluation with the weights is named as \textbf{Debiased Eva}. This can show how the leakage-neutral evaluation proposed in Section~\ref{sec:unbiased_learning} affect the evaluation results.
    \item \textbf{Synthetic dataset evaluation}. We evaluate the performance of models with the synthetic dataset introduced in Section~\ref{sec:fit_bias}. Given the prior probabilities of positive/negative classes fixed, a better model is supposed to give higher accuracy, and tended to be less impacted by the bias pattern.
    \item  \textbf{Cross-dataset evaluation}. We evaluate that how the models perform on other STS datasets, \emph{i.e.}, MSRP and SICK$_{\text{STS}}$. We use the entire datasets for evaluations. As the preparation strategies of different datasets are different, cross-dataset evaluations will not give additional rewards for the selection bias of QuoraQP. Although different datasets may have different contexts, a better model trained with QuoraQP is still supposed to perform better.
\end{itemize}

Among all the evaluation methods, using the testing set for evaluation without weights~(\emph{Biased Eva}) is biased, and we will show that the \emph{Debiased Eva} is more consistent with the unbiased synthetic dataset evaluation and cross-dataset evaluations.

\subsection{Experimental Results}
\label{sec:experimental_results}

\begin{table}[t!]
\begin{center}
\resizebox{0.485\textwidth}{!}{
\begin{tabular}{c|p{1.9cm}<{\centering}|p{2.375cm}<{\centering}}
\hline
\bf Method & \bf  Biased Eva & \bf  Debiased Eva \\
\hline
Majority & 50.00 & 51.62  \\
Leakage  & 79.63 & 54.40 \\
Biased Model & \textbf{83.97} & 78.76 \\
Debiased Model & 82.90 & \textbf{80.11} \\
\hline
\end{tabular}
}
\end{center}
\caption{\label{table:testset_eva} Evaluation Results with the testing set of QuoraQP. We report the accuracy scores and ``\%'' is omitted.}
\end{table}

\begin{table}[t!]
\begin{center}
\resizebox{0.485\textwidth}{!}{
\begin{tabular}{c|p{1.375cm}<{\centering}|p{1.2cm}<{\centering}|p{1.2cm}<{\centering}}
\hline
\bf Method & \bf Synthetic & \bf MSRP & \bf SICK$_{\text{STS}}$ \\
\hline
Biased Model & 89.46 & 51.94 & 64.95 \\
Debiased Model & \textbf{92.62} & \textbf{56.77} & \textbf{66.05} \\
\hline
\end{tabular}
}
\end{center}
\caption{\label{table:unbiased_eva} Evaluation Results with the synthetic dataset, MSRP and SICK$_{\text{STS}}$ dataset. We report the accuracy scores and ``\%'' is omitted.}
\end{table}

\begin{figure}[!t]
\includegraphics[width=7.2cm]{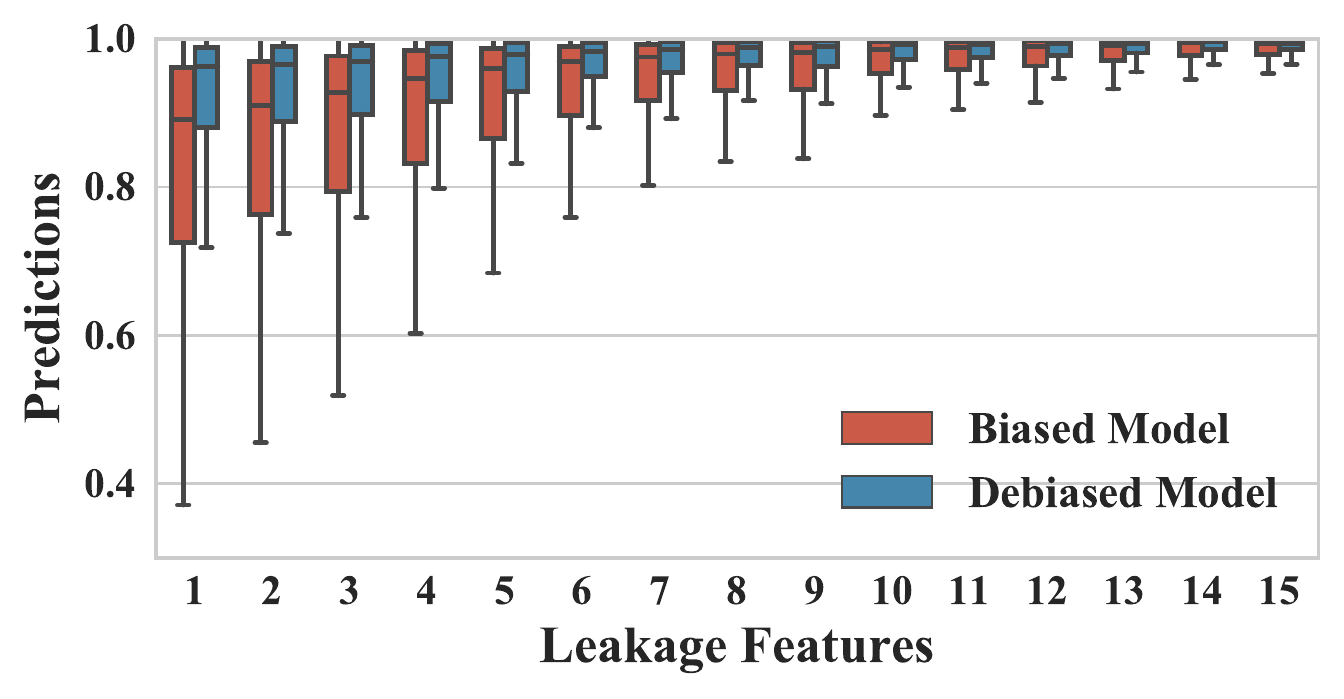}
\caption{Visualization of predicted scores by the Biased and Debiased Models versus the leakage feature. Red boxes represent the results by the Biased Model, and blue boxes represent the results by the Debiased Model. Best viewed in color.}
\label{fig:duplicated_preds_compare}
\end{figure}

The evaluation results on the testing set of QuoraQP are reported in Table~\ref{table:testset_eva}. From the accuracy of the method \emph{Leakage}, we can see that although the influence isn't completely eliminated, the evaluation result of Debiased Eva is less impacted by the bias pattern in the original distribution. This makes the results more reliable for evaluations. The reason why in the Leakage method the bias could not be completely eliminated is that we cannot estimate $P(S=y|l)$ perfectly. A minor error of $P(S=y|l)$ may result in a significant difference in the weight especially when the probability is close to zero, since the multiplicative inverse is used.

As for the Biased Model and the Debiased Model, we find that the Biased Model performs significantly better under the Biased Eva. This is the effect of fitting the bias pattern in addition to the semantic pattern, thus taking some extra advantage that cannot be generalized to real-life cases. On the other hand, under the Debiased Eva, we can find that the Debiased Model performs the best.

Table~\ref{table:unbiased_eva} reports the results on the datasets that are not biased to the leakage pattern of QuoraQP. We find that the Debiased Model significantly outperforms the Biased Model on all three datasets. This indicates that the Debiased Model better captures the true semantic similarities of the input sentences. We further visualize the predictions on the synthetic dataset in Figure~\ref{fig:duplicated_preds_compare}. As illustrated, the predictions are more neutral to the leakage feature.

From the experimental results, we can see that the proposed leakage-neutral training method is effective, as the Debiased Model performs significantly better with Synthetic dataset, MSRP and SICK, showing a better generalization strength. Moreover, the Debiased Eva gives results that are more consistent with the results on unbiased datasets, thus it can serve as a more reliable in-domain way to evaluate models trained with QuoraQP. As a conclusion, our constructed leakage-neutral distribution is more close to the real-world one compared with the biased distribution that is directly observed from the given datasets.

\section{Related Work}
\label{sec:related_work}

In this section, we summarize the related work and distinguish them from our contributions.

\paragraph{Inverse propensity score for debiasing}
Usually, the Inverse Propensity Score~(IPS) is used to reduce the selection bias~\citep{schonlau2009selection,d1998propensity}, where the propensity score~\citep{rosenbaum1983central} is the probability that a sample will be selected into the dataset. \citet{zadrozny2004learning} studies the learning and evaluating of classifiers under sample selection bias, while his focus was the ``missing-at-random''~(MAR)~\citep{little2014statistical} problem where the biasedness only depends on the feature vector $x$.

For NLSM datasets, the selection bias is ``not-missing-at-random''~(NMAR)~\citep{little2014statistical}, thus we cannot hope to estimate the true propensity scores directly as it requires the labels of unselected samples~\citep{zadrozny2004learning}. In this paper, we propose to fit a constructed leakage-neutral distribution, which could be achieved with only the selected samples that we can access.

\paragraph{Biasedness of datasets}
Although dataset bias is often mentioned, the research community is not putting sufficient attention to it compared with models and algorithms. \citet{torralba2011unbiased} studied the dataset bias for image recognition datasets, and categorize the bias into \emph{Selection Bias}, \emph{Capture Bias} and \emph{Negative Set Bias}. Selection bias is widely studied in the search ranking field as position bias~\citep{wang2016learning,wang2018position,joachims2017unbiased}. Usually the propensity scores are estimated through online Result Randomization~\citep{joachims2017unbiased}. \citet{liang2019additive} studied the biasedness for authentication, and proposed an additive adversarial learning for unbiased learning.

In the NLP field, \citet{minka2008selection} studied the selection bias in the LETOR datasets, and found that Reverse BM25 performs unreasonably well due to the selection procedure. \citet{dixon2018measuring} studied the potential unfairness for toxic comments classification due to unintended bias, and proposed methods to mitigate it by balancing the training dataset with additional data. \citet{gururangan2018annotation} and \citet{poliak2018hypothesis} found that in some NLI datasets, there is biasedness of specific linguistic phenomena, which makes it possible to classify the relationship of a pair of sentences, by only looking at one of them. \citet{sugawara2018makes} investigated what makes questions easier across recent 12 Machine Reading Comprehension~(MRC) datasets and the results suggest that one might overestimate recent advances in MRC.

In this paper, we study the selection bias embodied in the comparing relationships in NLSM datasets. To the best of our knowledge, this is the first study on this kind of selection bias.

\section{Conclusion}
\label{sec:conclusion}

In this paper, we take a close look at the selection bias of NLSM datasets and focus on the selection bias embodied in the comparing relationships of sentences. To mitigate the bias, we propose an easy-adopting method for leakage-neutral learning and evaluations. 

However, there is still much to do to form a clearer scope of this problem. For example, we still do not know the details of dataset preparations of many other NLSM datasets, and we can not say to what extent the assumptions in Section~\ref{sec:unbiased_learning} hold in QuoraQP and what is the relationship between the leakage-neutral distribution and the real-world distribution. We suggest for future NLSM datasets, the providers should pay more attention to this problem. Furthermore, they could reveal the more detailed strategy of sample selection, and might publish some official weights to eliminate the bias.

\bibliography{acl2019}
\bibliographystyle{acl_natbib}


\appendix

\section{Detailed Settings for the Experiments in Section~\ref{sec:leakage_in_datasets}}
\label{sec:appendix}

\subsection{Dataset Description}

We summarize the statistics of the datasets used in Section~\ref{sec:look_at_leakage} in Table~\ref{table:datasets}.

\begin{table}[h]
\begin{center}
\begin{tabular}{c|c|c|c}
\hline
\bf Dataset & \bf Training & \bf Testing & \bf \# classes \\
\hline
SNLI        & 549k & 10k & 3 \\
MultiNLI    & 393k & 10k & 3 \\
QuoraQP     & 384k & 10k & 2 \\
MSRP        & 4k   & 2k  & 2 \\
SICK        & 5k    & 5k    & 2/3 \\
ByteDance   & 256k & 32k & 3 \\
\hline
\end{tabular}
\end{center}
\caption{\label{table:datasets} Information about the datasets.}
\end{table}

For SICK, both \texttt{entailment\_label} and \texttt{relatedness\_score} are provided. We use the sentence pairs with \texttt{relatedness\_score} greater than $3.6$ as \texttt{duplicated}, and otherwise \texttt{not\_duplicated}. This threshold gives roughly 50\% of positive pairs and 50\% negative pairs.

For ByteDance, since no existing dataset partition is available, we randomly divide the dataset into a training set, a validation set, and a testing set in a ratio of 8:1:1. We use the sentences in English during our experiments. 

\subsection{Features Used in Unlexicalized}

We list the 15 features we used in method \textbf{Unlexicalized} in Section~\ref{sec:leakage_in_datasets}. We use 3 types of unlexicalized features~\citep{bowman2015large}:

\begin{itemize}\setlength{\itemsep}{0pt}
    \item The BLEU score of both sentences, using n-gram length from 1 to 4, which are totally 4 features.
    \item The length difference between the two sentences, as one real-valued feature.
    \item The number and percentage of overlap words between both sentences over all words and over just nouns, verbs, adjectives and adverbs, which are totally 10 features.
\end{itemize}

\subsection{Features Used in Advanced}

We list the features we used in method \textbf{Advanced} in Section~\ref{sec:leakage_in_datasets}. 
As mentioned above, if we use a node to represent a sentence and add an undirected edge if two sentences are compared in the dataset, the whole dataset can be viewed as a graph as illustrated in Figure~\ref{fig:toy_problem}. To classify the edges in the graph, we use 3 types of graph-based features:

\begin{itemize}\setlength{\itemsep}{0pt}
    \item The origin and extended leakage features: degrees of both nodes, number of 2-hop and 3-hop paths between the two nodes, number of 2-hop and 3-hop neighbors of both nodes, which are totally 8 features.
    \item The element-wise product and dot product of Deepwalk~\citep{perozzi2014deepwalk} embedding of the two nodes, all together as 65 features.
    \item The resource allocation index, Jaccard coefficient, preferential attachment score and  Adamic-Adar index~\citep{zhou2009predicting,liben2007link} of both two nodes, which are totally 4 features.
\end{itemize}

\section{Proof for the Theorems}

\subsection{Derivation of Equation~(\ref{eq})}
\label{sec:proof2}

Here we present the derivation of Equation~(\ref{eq}).

\begin{proof}
\begin{small}
\begin{displaymath}
\begin{split}
P_{\widehat{\mathscr{D}}}(Y & =1|l) \\
= & P(Y=1|S=Y,l)\\
= & \frac{P(Y=1,S=1|l)}{P(Y=1,S=1|l)+P(Y=0,S=0|l)} \\
= & \frac{P(Y=1|l)P(S=1|l)}{P(Y=1|l)P(S=1|l)+P(Y=0|l)P(S=0|l)} \\
= & \frac{P(Y=1)P(S=1|l)}{P(Y=1)P(S=1|l)+P(Y=0)P(S=0|l)} \text{.}
\end{split}
\end{displaymath}
\end{small}
By solving the above equation, we have the result in Equation~(\ref{eq}).
\end{proof}

\subsection{Proof of Theorem~\ref{theorem:unbiased}}
\label{sec:proof1}

Here we present the proof for Theorem~\ref{theorem:unbiased}, \emph{i.e.}, the unbiased expectation theorem.

\begin{proof}
\begin{small}
\begin{displaymath}
\begin{split}
& E_{x,y,l \sim \widehat{\mathscr{D}}} \Big[w\Delta\big(f(x,l), y\big) \Big] \\
= & \int \frac{P(S=Y)}{P(S=y|l)} \Delta(f(x,l), y) dP_{\widehat{\mathscr{D}}}(x,y,l) \\
= & \int \Delta(f(x,l), y) \frac{P(S=Y)}{P(S=y|l)} dP(x,y,l|S=Y) \\
= & \int \Delta(f(x,l), y) \frac{P(S=Y)}{P(S=y|l)} \frac{P(S=y|x,y,l)dP(x,y,l)}{P(S=Y)} \\
= & \int \Delta(f(x,l), y) dP(x, y, l) \\
= & E_{x, y, l \sim \mathscr{D}} \Big[ \Delta(f(x,l), y) \Big] \text{.}
\end{split}
\end{displaymath}
\end{small}
\end{proof}

As illustrated above, by adding specific weights to the samples, we can obtain the loss unbiased to the leakage neutral distribution $\mathscr{D}$. The unbiased loss can be used for both training and evaluation.

\end{document}